\documentclass[11pt,letterpaper]{article}
\usepackage{emnlp2017}
\usepackage{times}
\usepackage{latexsym}
\usepackage{graphicx}
\usepackage{mathtools}
\usepackage{subcaption}
\usepackage{caption}
\usepackage{xcolor}
\usepackage{multirow}
\usepackage{amssymb}
\usepackage{bm}
\usepackage{wrapfig}
\emnlpfinalcopy
\def\emnlppaperid{4}

\title{Shakespearizing Modern Language Using Copy-Enriched Sequence-to-Sequence Models}

 \author{Harsh Jhamtani \thanks{* denotes equal contribution}, Varun Gangal  \footnotemark[1], Eduard Hovy, Eric Nyberg \\
         Language Technologies Institute \\ Carnegie Mellon University \\ {\tt \{jharsh,vgangal,hovy,ehn\}@cs.cmu.edu } }

\date{}

\begin{document}

\maketitle

\begin{abstract}
Variations in writing styles are commonly used to adapt the content to a specific context, audience, or purpose. However, applying stylistic variations is still largely a manual process, and there have been little efforts towards automating it. In this paper we explore automated methods to transform text from modern English to Shakespearean English using an end to end trainable neural model with pointers to enable copy action. To tackle limited amount of parallel data, we pre-train embeddings of words by leveraging external dictionaries mapping Shakespearean words to modern English words as well as additional text. Our methods are able to get a BLEU score of $31+$, an improvement of $\approx6$ points over the strongest baseline. We publicly release our code to foster further research in this area. \footnote{https://github.com/harsh19/Shakespearizing-Modern-English}

\end{abstract}

\section{Introduction}

Text is often morphed using a variety of lexical and grammatical transformations, adjusting the degree of formality, usage of catchy phrases, and other such stylistic changes to make it more appealing.
Moreover, different text styles appeal to different user segments \cite{saha2015automated} \cite{kitis1997ads} \cite{schwartz2013personality}. 
Thus there is a need to effectively adapt text to different styles. However, manually transforming text to a desired style can be a tedious process.

\begin{table}[!htbp]
\centering
\tiny
\addtolength{\tabcolsep}{-4pt}
\begin{tabular}{|l|l|l| }
\hline 
No & Type  & Text \\ \hline \hline
\multirow{3}{*}{1} &   \textsc{Modern} & Oh my, my bones ache so much \\
&  \textsc{Original} & Fie, how my bones ache ! \\
& \textsc{Copy} & fie, how my bones ache ! \\  
& \textsc{SimpleS2S} & you'll be, sir, what the bones are tired . \\
& \textsc{Stat} & Oh my, my bones ache so much . \\ \hline \hline
\multirow{3}{*}{2} &  \textsc{Modern} & I am in a rush . \\
&  \textsc{Original} & I stand on sudden haste .  \\
& \textsc{Copy} & i stand on sudden haste . \\  
& \textsc{SimpleS2S} & i'm stand right here . \\
& \textsc{Stat} & I am in a Fly \\ \hline \hline
\multirow{3}{*}{3} 
&  \textsc{Modern} & Give my compliments to your lady  \\
&  \textsc{Original} & Commend me to thy lady \\
& \textsc{Copy} & commend me to your lady \\  
& \textsc{SimpleS2S} & give my regards to your lady \\
& \textsc{Stat} & give my praises to your lady \\ \hline \hline

\multirow{3}{*}{4}
&  \textsc{Modern} & Showing mercy by pardoning killers only causes more murders .  \\
 &  \textsc{Original} & Mercy but murders, pardoning those that kill . \\
& \textsc{Copy} & mercy but murders, those those who kill us . \\  
& \textsc{SimpleS2S} & but except the murders to those murders to kill you . \\
& \textsc{Stat} & of mercy by pardoning killers causes more dire. \\ \hline \hline
\multirow{3}{*}{5} 
& \textsc{Modern} & Holy Saint Francis, this is a drastic change !  \\
&  \textsc{Original} & Holy Saint Francis, what a change is here ! \\
& \textsc{Copy} & holy saint francis, what a change is here ! \\  
& \textsc{SimpleS2S} & it's the holy flute, what's the changed ! \\
& \textsc{Stat} & Holy Saint Francis, this is a drastic change ! \\ \hline \hline
\multirow{3}{*}{6} 
&  \textsc{Modern} & was that my father who left here in such a hurry ?  \\
&  \textsc{Original} & Was that my father that went hence so fast ?  \\
& \textsc{Copy} & was that my father that went went so fast ? \\  
& \textsc{SimpleS2S} & was that my father was so that ? \\
& \textsc{Stat} & was that my father that left here in such a haste ? \\ 
\hline \hline
\multirow{3}{*}{7} 
& \textsc{Modern} & Give me one kiss and I'll go down .  \\
&  \textsc{Original} & One kiss, and I'll descend .  \\
& \textsc{Copy} & one kiss me, and I'll descend . \\  
& \textsc{SimpleS2S} & one kiss,and I come down . \\
& \textsc{Stat} & Give me a kiss, and I'll go down . \\ \hline \hline
\multirow{3}{*}{8} 
&  \textsc{Modern} &  then the window lets day in, and life goes out the window .  \\
&  \textsc{Original} & Then, window, let day in and life out .  \\
& \textsc{Copy} & then, window out, and day life . \\  
& \textsc{SimpleS2S} & then she is just a life of life, let me life out of life . \\
& \textsc{Stat} & then the window will let day in, and life out . \\ \hline
\hline
\end{tabular}
\caption{Examples from dataset showing modern paraphrases (\textsc{Modern}) of few sentences from Shakespeare's plays (\textsc{Original}). We also show transformation of modern text to Shakespearean text from our models (\textsc{Copy}, \textsc{SimpleS2S} and \textsc{Stat}).}
\label{tab:intro}
\end{table}


There have been increased efforts towards machine assisted text content creation and editing through automated methods for summarization \cite{rush2015neural} , brand naming \cite{namification}, text expansion \cite{srinivasan2017corpus}, etc. However, there is a dearth of automated solutions for adapting text quickly to different styles. We consider the problem of transforming text written in modern English text to Shakepearean style English. For the sake of brevity and clarity of exposition, we henceforth refer to the \textit{Shakespearean} sentences/side as \textit{Original} and the modern English paraphrases as \textit{Modern}.

Unlike traditional domain or style transfer, our task is made more challenging by the fact that the two styles employ diachronically disparate registers of English - one style uses the contemporary language while the other uses \textit{Early Modern English \footnote{\url{https://en.wikipedia.org/wiki/Early_Modern_English}}} from the \textit{Elizabethan Era} (1558-1603). Although \textit{Early Modern English} is not classified as a different language (unlike \textit{Old English} and \textit{Middle English}), it does have novel words (\textit{acknown} and \textit{belike}), novel grammatical constructions (two \textit{second person} forms - \textit{thou} (informal) and \textit{you} (formal) \cite{brown1960pronouns}), semantically drifted senses (e.g \textit{fetches} is a synonym of \textit{excuses}) and non-standard orthography \cite{rayson2007tagging}. 
Additionally, there is a domain difference since the Shakespearean play sentences are from a dramatic screenplay whereas the \emph{parallel} modern English sentences are meant to be simplified explanation for high-school students. 

Prior works in this field leverage a language model for the target style, achieving transformation either using phrase tables \cite{xu2012paraphrasing}, or by inserting relevant adjectives and adverbs \cite{saha2015automated}. Such works have limited scope in the type of transformations that can be achieved. Moreover, statistical and rule MT based systems do not provide a direct mechanism to a) share word representation information between source and target sides b) incorporating constraints between words into word representations in end-to-end fashion. Neural sequence-to-sequence models, on the other hand, provide such flexibility. 


Our main contributions are as follows:
\begin{itemize}
    \item We use a sentence level sequence to sequence neural model with a pointer network component to enable direct copying of words from input. We demonstrate that this method performs much better than prior phrase translation based approaches for transforming \textit{Modern} English text to \emph{Shakespearean} English. 
    \item We leverage a dictionary providing mapping between Shakespearean words and modern English words to retrofit pre-trained word embeddings. Incorporating this extra information enables our model to perform well in spite of small size of parallel data. 
\end{itemize}

Rest of the paper is organized as follows. We first provide a brief analysis of our dataset in  (\S\ref{sec:Dataset}). We then elaborate on details of our methods in  (\S\ref{sec:Method}, \S\ref{sec:Method2}, \S\ref{sec:Method3}, \S\ref{sec:Method4}). We then discuss experimental setup and baselines in (\S\ref{sec:Experiments}). Thereafter, we discuss the results and observations in (\S \ref{sec:Results}). We conclude with discussions on related work (\S \ref{sec:RelatedWord}) and future directions (\S \ref{sec:Conclusion}).

\section{Dataset} \label{sec:Dataset}

Our dataset is a collection of line-by-line modern paraphrases for 16 of Shakespeare's 36 plays (\textit{Antony \& Cleopatra}, \textit{As You Like It}, \textit{Comedy of Errors}, \textit{Hamlet}, \textit{Henry V} etc) from the educational site \textit{Sparknotes}\footnote{\url{www.sparknotes.com}}.
This dataset was compiled by Xu et al. \shortcite{xu2014data,xu2012paraphrasing} and is freely available on github.\footnote{ \url{http://tinyurl.com/ycdd3v6h}}
14 plays covering 18,395 sentences form the training data split. We kept 1218 sentences from the play \emph{Twelfth Night} as validation data set. The last play, \emph{Romeo and Juliet}, comprising of 1462 sentences, forms the test set.

\begin{table}
\centering
\scriptsize
\addtolength{\tabcolsep}{-2pt}
\begin{tabular}{|l|l|l| }
\hline 
{} & \textit{Original}  & \textit{Modern} \\ \hline \hline
$\#$ Word Tokens & 217K & 200K \\ \hline
$\#$ Word Types & 12.39K  & 10.05K \\ \hline
Average Sentence Length & 11.81  & 10.91 \\ \hline
Entropy (Type.Dist) & 6.15 & 6.06 \\ \hline
$\cap$ Word Types       & \multicolumn{2}{|l|}{6.33K} \\
\hline \hline
\end{tabular}
\caption{Dataset Statistics}
\label{tab:profile} 
\end{table}

\subsection{Examples}
Table \ref{tab:intro} shows some parallel pairs from the test split of our data, along with the corresponding target outputs from some of our models. \textit{Copy} and \textit{SimpleS2S} refer to our best performing attentional S2S models with and without a \textit{Copy} component respectively. \textit{Stat} refers to the best statistical machine translation baseline using off-the-shelf GIZA++ aligner and MOSES. We can see through many of the examples how direct copying from the source side helps the \textit{Copy} generates better outputs than the \textit{SimpleS2S}. The approaches are described in greater detail in (\S\ref{sec:Method}) and (\S\ref{sec:Experiments}).

\subsection{Analysis}
Table \ref{tab:profile} shows some statistics from the training split of the dataset. In general, the \textit{Original} side has longer sentences and a larger vocabulary. The slightly higher entropy of the \textit{Original} side's frequency distribution indicates that the frequencies are more spread out over words. Intuitively, the large number of shared word types indicates that sharing the representation between \textit{Original} and \textit{Modern} sides could provide some benefit.

\section{Method Overview} \label{sec:Method}

\begin{figure*}
\centering
\includegraphics[scale=0.50]{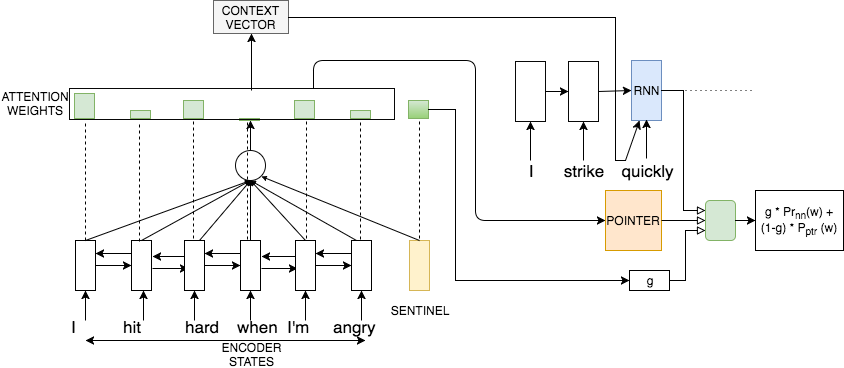}
\caption{Depiction of our overall architecture (showing decoder step 3). Attention weights are computed using previous decoder hidden state $h_2$, encoder representations, and sentinel vector. Attention weights are shared by decoder RNN and pointer models. The final probability distribution over vocabulary comes from both the decoder RNN and the pointer network. Similar formulation is used over all decoder steps }
\label{fig:architecture}
\end{figure*}

Overall architecture of the system is shown in Figure \ref{fig:architecture}. We use a bidirectional LSTM to encode the input modern English sentence. 
Our decoder side model is a mixture model of RNN module amd pointer network module. The two individual modules share the attentions weights over encoder states, although it is not necessary to do so. The decoder RNN predicts probability distribution of next word over the vocabulary, while pointer model predicts probability distribution over words in input. The two probabilities undergo a weighted addition, the weights themselves computed based on previous decoder hidden state and the encoder outputs. 


Let $\mathbf{x}, \mathbf{y}$ be the some input - output sentence pair in the dataset. Both input $\mathbf{x}$ as well as output $\mathbf{y}$ are sequence of tokens. $\mathbf{x} = \mathbf{x}_1 \mathbf{x}_2 ... \mathbf{x}_{T_{enc}}$, where $T_{enc}$ represents the length of the input sequence $\mathbf{x}$. Similarly, $\mathbf{y} = \mathbf{y}_1 \mathbf{y}_2 ... \mathbf{y}_{T_{dec}} $. Each of $\mathbf{x}_i $, $\mathbf{y}_j $ is a token from the vocabulary.

\section{Token embeddings} \label{sec:Method2}
Each token in vocabulary is represented by a $M$ dimensional embedding vector. Let vocabulary $V$ be the union of modern English and Shakepearean vocabularies i.e. $V = V_{shakespeare} \cup V_{modern}$.  $E_{enc}$ and $E_{dec}$ represent the embedding matrices used by encoder and decoder respectively ( $  E_{enc}, E_{dec} \in \mathbb{R}^{|V| \times M}  $ ). We consider union of the vocabularies for both input and output embeddings because many of the tokens are common in two vocabularies, and in the best performing setting we share embeddings between encoder and decoder models.
Let $E_{enc}(t)$, represent encoder side embeddings of some token $t$. For some input sequence $\mathbf{x}$, $E_{enc}(\mathbf{x})$ is given as $( E_{enc}(\mathbf{x}_1), E_{enc}(\mathbf{x}_2), ... ) $.

\subsection{Pretraining of embeddings}
Learning token embeddings from scratch in an end-to-end fashion along with the model greatly increases the number of parameters. 
To mitigate this, we consider pretraining of the token embeddings. 
We pretrain our embeddings on all training sentences. We also experiment with adding additional data from PTB \cite{marcus1993building} for better learning of embeddings. 
Additionally we leverage a dictionary mapping tokens from Shakespearean English to modern English.

We consider four distinct strategies to train the embeddings. In the cases where we use external text data, we first train the embeddings using both the external data and training data, and then for the same number of iterations on training data alone, to ensure adaptation. Note that we do not directly use off-the-shelf pretrained embeddings such as \textit{Glove} \cite{pennington2014glove} and \textit{Word2Vec} \cite{mikolov2013efficient} since we need to learn embeddings for novel word forms (and also different word senses for extant word forms) on the \textit{Original} side.

\subsubsection{Plain}
This method is the simplest pre-training method. Here, we do not use any additional data, and train word embeddings are trained on the union of \textit{Modern} and \textit{Original} sentences. 

\subsubsection{PlainExt}
In this method, we add all the sentences from the external text source (\textit{PTB}) in addition to sentences in training split of our data. 

\subsubsection{Retro}
We leverage a dictionary $L$ of approximate \textit{Original} $\rightarrow$ \textit{Modern} word pairs \cite{xu2012paraphrasing,xu2014data}, crawled from \url{shakespeare-words.com}, a source distinct from Sparknotes. We explicitly add the two \textit{2nd persons} and their corresponding forms (thy, thou, thyself etc) which are very frequent but not present in $L$.
The final dictionary we use has 1524 pairs. Faruqui et al \shortcite{faruqui2014retrofitting} proposed a \emph{retrofitting} method to update a set of word embeddings to incorporate pairwise similarity constraints. Given a set of embeddings $p_i \in P$, a vocabulary $V$, and a set $C$ of pairwise constraints $(i,j)$ between words, retrofitting tries to learn a new set of embeddings $q_i \in Q$ to minimize the following objective:
\begin{center}
\footnotesize
\begin{align}
    f(Q) & = \delta \sum_{i=1}^{i=|V|} {(p_i-q_i)}^2 + \omega \sum_{(i,j) \in C} {(q_i-q_j)}^2
\end{align}
\normalsize
\end{center}
We use their off-the-shelf 
implementation \footnote{\url{github.com/mfaruqui/retrofitting}} to encode the dictionary constraints into our pretrained embeddings, setting $C=L$ and using suggested default hyperparameters for $\delta$, $\omega$ and number of iterations.

\subsubsection{RetroExt}
This method is similar to \emph{Retro}, except that we use sentences from the external data (\textit{PTB}) in addition to training sentences.

We use \textbf{None} to represent the settings where we do not pretrain the embeddings.


\subsection{Fixed embeddings}
Fine-tuning pre-trained embeddings for a given task may lead to \emph{overfitting}, especially in scenarios with small amount of supervised data for the task \cite{madhyastha2015mapping}. This is because embeddings for only a fraction of vocabulary items get updated,  leaving the embeddings unchanged for many vocabulary items. To avoid this, we consider fixed embeddings pretrained as per procedures described earlier. While reporting  results  in Section (\S \ref{sec:Results}), we separately report results for fixed (\emph{FIXED}) and trainable (\emph{VAR}) embeddings, and observe that keeping embeddings fixed leads to better performance. 

\section{Method Description} \label{sec:Method3}
In this section we give details of the various modules in the proposed neural model. 

\subsection{Encoder model}
 Let $\overrightarrow{LSTM_{enc}}$ and $\overleftarrow{LSTM_{enc}}$ represent the forward and reverse encoder. $\mathbf{h}^{\overrightarrow{enc}}_{t}$ represent hidden state of encoder model at step $t$ ($\mathbf{h}^{\overrightarrow{enc}}_{t} \in \mathbb{R}^{H} $).
The following equations describe the model:
\begin{center}
\footnotesize
\begin{align}
    \mathbf{h}^{\overrightarrow{enc}}_0 &=\overrightarrow{0},  \mathbf{h}^{\overleftarrow{enc}}_{|x|} = \overrightarrow{0} \\
    \mathbf{h}^{\overrightarrow{enc}}_{t} &= \overrightarrow{LSTM_{enc}}(\mathbf{h}^{enc}_{t-1},{E_{enc}}({\mathbf{x}_{t}})) \\
    \mathbf{h}^{\overleftarrow{enc}}_{t} &= \overleftarrow{LSTM_{enc}}(\mathbf{h}^{enc}_{t+1},{E_{enc}}({x_{t}})) \\
    \mathbf{h}^{enc}_{t} &= \mathbf{h}^{\overrightarrow{enc}}_{t} + \mathbf{h}^{\overleftarrow{enc}}_{t} 
\end{align}
\normalsize
\end{center}
We use addition to combine the forward and backward encoder states, rather than concatenation which is standardly used, since it doesn't add extra parameters, which is important in a low-data scenario such as ours. 


\subsection{Attention}

Let $\mathbf{h}_t^{dec}$ represent the hidden state of the decoder LSTM at step $t$. Let $E_{dec}(\mathbf{y}_{t-1})$ represent the decoder side embeddings of previous step output. We use special $START$ symbol at $t=1$. 

We first compute a query vector, which is a linear transformation of $\mathbf{h}_{t-1}^{dec}$. A sentinel vector $\mathbf{s} \in \mathbb{R}^H$ is concatenated with the encoder states to create $F_{att} \in \mathbb{R} ^ { (T_{enc}+1) \times H } $, where $T_{enc}$ represents the number of tokens in encoder input sequence $\mathbf{x}$. A normalized attention weight vector $\boldsymbol{\alpha}^{norm}$ is computed. The value $g$, which corresponds to attention weight over sentinel vector, represents the weight given to the decoder RNN module while computing output probabilties.
\begin{center}
\footnotesize
\begin{align}
  \mathbf{q} &= \mathbf{h}_{t-1}^{dec} \,W_{q}   &&  W_{q} \in \mathbb{R}^{H \times H} \\
  F_{att} &= concat( \mathbf{h}^{enc}_{1..T_{enc}}, \mathbf{s} )   && F_{att} \in \mathbb{R}^{(T_{enc}+1) \times H} \\
  \boldsymbol{\alpha}_i &= \sum_{j=1}^{H}( tanh(F_{att}^{(ij)} \, \mathbf{q}_j) ) + \mathbf{b}_i   && \boldsymbol{\alpha}_i, \mathbf{b}_i \in \mathbb{R} \\
  \boldsymbol{\alpha}^{norm} &= softmax(\boldsymbol{\alpha}) && \boldsymbol{\alpha}^{norm} \in \mathbb{R}^{T_{enc}+1} \\
  \boldsymbol{\beta} &= \boldsymbol{\alpha}^{norm}_{1,2,...,T_{enc}} && \boldsymbol{\beta} \in \mathbb{R}^{T_{enc}} \\
  g &= \boldsymbol{\alpha}^{norm}_{T_{enc}+1} && g \in \mathbb{R}
\end{align}
\normalsize
\end{center}


\subsection{Pointer model}

As pointed out earlier, a pair of corresponding \textit{Original} and \textit{Modern} sentences have significant vocabulary overlap. Moreover, there are lot of proper nouns and rare words which might not be predicted by a sequence to sequence model. To rectify this, pointer networks have been used to enable copying of tokens from input directly \cite{merity2016pointer}. The pointer module provides location based attention, and output probability distribution due to pointer network module can be expressed as follows:
\begin{center}
\footnotesize
\begin{align}
P_{t}^{PTR}(w) &= \sum_{\mathbf{x}_j=w}( \boldsymbol{\beta}_j )
\end{align}
\normalsize
\end{center}

\subsection{Decoder RNN}

Summation of encoder states weighed by corresponding attention weights yields context vector. 
Output probabilities over vocabulary as per the decoder LSTM module are computed as follows:

\begin{center}
\footnotesize
\begin{align}
\mathbf{c}_t &= \sum_{i=1}^{T_{enc}} \boldsymbol{\beta}_i \, \mathbf{h}^{enc}_i  \\
\mathbf{h}^{dec}_{t} &= LSTM(\mathbf{h}^{dec}_{t-1},[\text{concat}({E_{dec}}({\mathbf{y}_{t-1}}),\mathbf{c}_{t})])  \\
P_{t}^{LSTM} &=\text{softmax}(W_{out}[\text{concat}(\mathbf{h}^{dec}_{t}, \mathbf{c}_{t})] + \mathbf{b}^{out}) 
\end{align}
\normalsize
\end{center}
During training, we feed the ground truth for $\mathbf{y}_{t-1}$, whereas while making predictions on test data, predicted output from previous step is used instead.

\subsection{Output prediction}
Output probability of a token $w$ at step $t$ is a weighted sum of probabilities from decoder LSTM model and pointer model given as follows:
\begin{center}
\footnotesize
\begin{align}
P_{t}(w) &= g \times P_{t}^{LSTM}(w) + (1-g) \times P_{t}^{PTR}(w) 
\end{align}
\normalsize
\end{center}

$P_{t}^{PTR}(w)$ takes a non-zero value only if $w$ occurs in input sequence, otherwise it is $0$. 
Forcing $g=0$ would correspond to not having a \textit{Copy} component, reducing the model to a plain attentional S2S model, which we refer to as a \textit{SimpleS2S} model. 
%
%
%
%

\section{Loss functions} \label{sec:Method4}

Cross entropy loss is used to train the model.
For a data point $( \mathbf{x}, \mathbf{y} ) \in \mathcal{D}$ and predicted probability distributions $P_t\,(w)$ over the different words $w \in \mathbf{V}$ for each time step $t \in \{1,\ldots,T_{dec}\}$, the loss is given by
\begin{center}
\small
\begin{align}
  - \sum_{t=1}^{T_{dec}} \,\log p\,\bigl(P_t\,(\mathbf{y}_t) \bigr)
\end{align}
\normalsize
\end{center}

\textbf{Sentinel Loss (\textsc{SL}):}
Following from work by \cite{merity2016pointer}, we consider additional sentinel loss. This loss function can be considered as a form of \emph{supervised attention}. Sentinel loss is given as follows:
\begin{center}
\small
\begin{align}
  - \sum_{t=1}^{T_{dec}} \,\log ( g^{(t)} + \sum_{x_j=y_t}( \beta^{(t)}_j ) )
\normalsize
\end{align}
\end{center}

We report the results demonstrating the impact of including the sentinel loss function (\textsc{+SL}).

%
%
\section{Experiments} \label{sec:Experiments}
In this section we describe the experimental setup and evaluation criteria used. 

\subsection{Preprocessing}
We lowercase sentences and then use NLTK's PUNKT tokenizer to tokenize all sentences. The \textit{Original} side has certain characters like \ae which are not extant in today's language. We map these characters to the closest equivalent character(s) used today (e.g \ae $\rightarrow$ ae)

\subsection{Baseline Methods}
\subsubsection{As-it-is}
Since both source and target side are English, just replicating the input on the target side is a valid and competitive baseline, with a BLEU of $21+$.

\subsubsection{Dictionary}
Xu et al. \shortcite{xu2012paraphrasing} provide a dictionary mapping between large number of Shakespearean and modern English words. 
We augment this dictionary with pairs corresponding to the $2$nd person thou (\textit{thou}, \textit{thy}, \textit{thyself}) since these common tokens were not present. 

Directly using this dictionary to perform word-by-word replacement is another admittable baseline. As was noted by Xu et al. \shortcite{xu2012paraphrasing},  this baseline actually performs worse than \textit{As-it-is}. This could be due to its performing aggressive replacement without regard for word context. Moreover, a dictionary cannot easily capture one-to-many mappings as well as long-range dependencies \footnote{thou-thyself and you-yourself}.

\subsubsection{Off-the-shelf SMT}
To train statistical machine translation (\textit{SMT}) baselines, we use publicly available open-source toolkit MOSES \cite{koehn2007moses}, along with the GIZA++  word aligner \cite{och2003minimum}, as was done in \cite{xu2012paraphrasing}. For training the target-side LM component, we use the \textit{lmplz} toolkit within MOSES to train a 4-gram LM. We also use \textit{MERT}  \cite{och2003minimum}, available as part of MOSES, to tune on the validation set.

For fairness of comparison, it is necessary to use the pairwise dictionary and \textit{PTB} while training the SMT models as well 
- the most obvious way for this is to use the dictionary and \textit{PTB} as additional training data for the alignment component and the target-side LM respectively. We experiment with several SMT models, ablating for the use of both \textit{PTB} and dictionary. 
In \ref{sec:Results}, we only report the performance of the best of these approaches. 


\subsection{Evaluation}
Our primary evaluation metric is \emph{BLEU} \cite{papineni2002bleu} . We compute \emph{BLEU} using the freely available and very widely used perl script\footnote{\url{http://tinyurl.com/yben45gm}} from the MOSES decoder.

We also report \emph{PINC} \cite{chen2011collecting}, which originates from paraphrase evaluation literature and evaluates how much the target side paraphrases resemble the source side. Given a source sentence $s$ and a target side paraphrase $c$ generated by the system, \emph{PINC(s,c)} is defined as
\begin{center}
\scriptsize
\begin{align*}
    PINC(s,c)&=1-\frac{1}{N} \sum_{n=1}^{n=N} \frac{|Ngram(c,n) \cap Ngram(s,n)|}{|Ngram(c,n)|}
\end{align*}
\normalsize
\end{center}
where $Ngram(x,n)$ denotes the set of n-grams of length $n$ in sentence $x$, and $N$ is the maximum length of ngram considered. We set $N=4$. Higher the \textit{PINC}, greater the novelty of paraphrases generated by the system. Note, however, that PINC does not measure fluency of generated paraphrases.

\subsection{Training and Parameters}
We use a minibatch-size of $32$ and the \textit{ADAM} optimizer \cite{kingma2014adam} with learning rate $0.001$,  momentum parameters $0.9$ and $0.999$, and $\epsilon=10^{-8}$. 
All our implementations are written in Python using Tensorflow 1.1.0 framework. 

For every model, we experimented with two configurations of embedding and LSTM size - $S$ (128-128), $ME$ (192-192) and $L$ (256-256). Across models, we find that the $ME$ configuration performs better in terms of highest validation BLEU.  We also find that larger configurations (384-384 \& 512-512) fail to converge or perform very poorly \footnote{This is expected given the small parallel data}. Here, we report results only for the $ME$ configuration for all the models.
For all our models, we picked the best saved model over 15 epochs which has the highest validation BLEU. 

\subsection{Decoding}
At test-time we use greedy decoding to find the most likely target sentence\footnote{Empirically, we observed that beam search does not give improvements for our task}. We also experiment with a post-processing strategy which replaces \emph{UNKs} in the target output with the highest aligned (maximum attention) source word. We find that this gives a small jump in \emph{BLEU} of about 0.1-0.2 for all neural models \footnote{Since effect is small and uniform, we report BLEU before post-processing in Table \ref{tab:knightExp} }. Our best model, for instance, gets a jump of 0.14 to reach a BLEU of \emph{\textbf{31.26}} from 31.12.

\section{Results} \label{sec:Results}

%


The results in Table \ref{tab:knightExp} confirm most of our hypotheses about the right architecture for this task.
\begin{itemize}
    \item \textbf{Copy component}: We can observe from Table \ref{tab:knightExp} that the various \emph{Copy} models each outperform their \emph{SimpleS2S} counterparts by atleast 7-8 BLEU points.
    \item \textbf{Retrofitting dictionary constraints}: The \emph{Retro} configurations generally outperform their corresponding \emph{Plain} configurations. For instance, our best configuration \emph{Copy.Yes.RetroExtFixed} gets a better BLEU than \emph{Copy.Yes.PlainExtFixed} by a margin of atleast 11.
    \item \textbf{Sharing Embeddings}: Sharing source and target side embeddings benefits all the \emph{Retro} configurations, although it slightly deteriorates performance (about 1 BLEU point) for some of the \emph{Plain} configurations. 
    \item \textbf{Fixing Embeddings}: \emph{Fixed} configurations always perform better than corresponding \emph{Var} ones (save some exceptions). For instance, \emph{Copy.Yes.RetroExtFixed} get a BLEU of 31.12 compared to 20.95 for \emph{Copy.Yes.RetroExtVar}. Due to fixing embeddings, the former has just half as many parameters as the latter (5.25M vs 9.40M)
    \item \textbf{Effect of External Data}: Pretraining with external data \emph{Ext} works well along with retrofitting \emph{Retro}. For instance, \emph{Copy.Yes.RetroExtFixed} gets a BLEU improvement of 2+ points over \emph{Copy.Yes.RetroFixed}
    \item \textbf{Effect of Pretraining}: For the \emph{SimpleS2S} models, pre-training adversely affects BLEU. However, for the \emph{Copy} models, pre-training leads to improvement in BLEU. The simplest pretrained \emph{Copy} model, \emph{Copy.No.PlainVar} has a BLEU score 1.8 higher than \emph{Copy.No.NoneVar}.
    \item \textbf{PINC scores}: All the neural models have higher PINC scores than the statistical and dictionary approaches, which indicate that the target sentences produced differ more from the source sentences than those produced by these approaches.
    \item \textbf{Sentinel Loss:} Adding the sentinel loss does not have any significant effect, and ends up reducing BLEU by a point or two, as seen with the \emph{Copy+SL} configurations.
\end{itemize}

\subsection{Qualitative Analysis}

\begin{figure}
\captionsetup[subfigure]{labelformat=empty}
\captionsetup[subfigure]{justification=centering}
\begin{subfigure}{0.2\textwidth}
\includegraphics[scale=0.20]{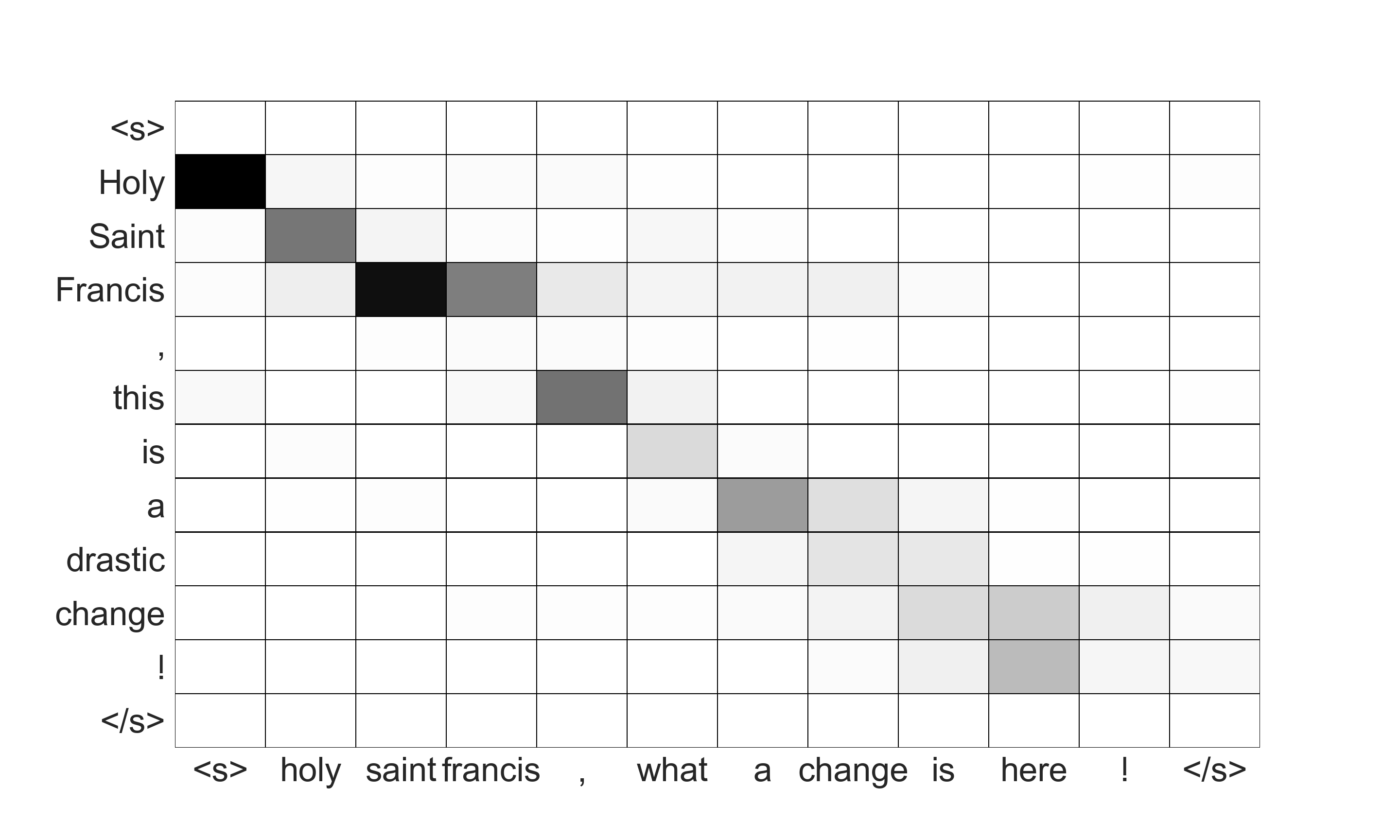}
\end{subfigure} \hspace{0.15\textwidth}
\begin{subfigure}{0.2\textwidth}
\centering
\includegraphics[scale=0.20]{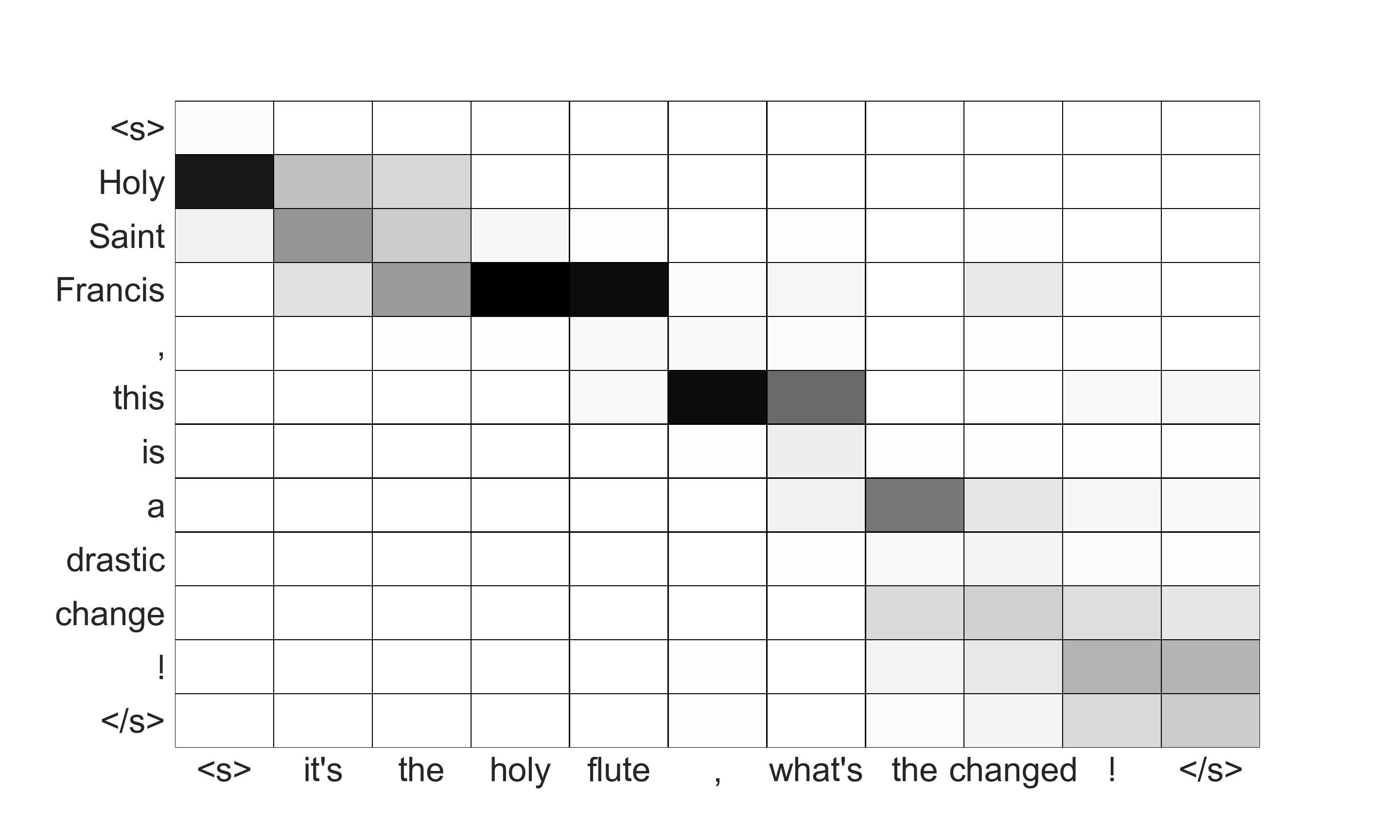}
\end{subfigure}
\caption{Attention matrices from a \emph{Copy} (top) and a \textit{simple S2S} (bottom) model respectively on the input sentence \textit{``Holy Saint Francis, this is a drastic change!"} . \textbf{$<s>$} and \textbf{$</s>$} are  start and stop characters. Darker cells are higher-valued.}
\textbf{\label{fig:attention}}
\end{figure}

Figure \ref{fig:attention} shows the attention matrices from our best \textit{Copy} model (\emph{Copy.Yes.RetroExtFixed}) and our best \textit{SimpleS2S} model (\emph{SimpleS2S.Yes.Retrofixed}) respectively for the same input test sentence. Without an explicit \textit{Copy} component, the \textit{SimpleS2S} model cannot predict the words \textit{saint} and \textit{francis}, and drifts off after predicting incorrect word \textit{flute}.

\begin{table}[!htbp]
\centering
\scriptsize
\addtolength{\tabcolsep}{-2pt}
\begin{tabular}{|l|l|l|l| }
\hline 
Model & Sh  & Init  & BLEU (PINC) \\ \hline \hline
\textsc{As-it-is}  & {-} & {-}  &  {21.13} (0.0)  \\ \hline
\textsc{Dictionary}  & {-} & {-}  &  {17.00} (26.64)  \\ \hline
\textsc{Stat}   & {-} & {-}  &  \textbf{24.39} (32.30)    \\ \hline
\multirow{10}{*}{\textsc{SimpleS2S}} &  $\times$ & $NoneVar$ & 11.66 (85.61) \\
&  $\times$ & $PlainVar$ & 9.27 (86.52) \\
 & $\times$ & $PlainExtVar$  & 8.73 (87.17) \\ 
 & $\times$ & $RetroVar$ &  10.57 (85.06) \\ 
& $\times$ & $RetroExtVar$  & 10.26 (83.83) \\ 
& $\checkmark$ & $NoneVar$ &  11.17 (84.91) \\
 & $\checkmark$ & $PlainVar$ &  8.78 (85.57) \\
 & $\checkmark$ & $PlainFixed$ &  8.73 (89.19)\\
 & $\checkmark$ & $PlainExtVar$  & 8.59 (86.04) \\
 & $\checkmark$ & $PlainExtFixed$  & 8.59 (89.16) \\
 & $\checkmark$ & $RetroVar$ &  10.86 (85.58) \\
 & $\checkmark$ & $RetroFixed$ &  11.36 (85.07) \\
 & $\checkmark$ & $RetroExtVar$  & 11.25 (83.56) \\
 & $\checkmark$ & $RetroExtFixed$  & \textbf{10.86} (88.80) \\  \hline
\multirow{6}{*}{\textsc{Copy}} & $\times$ & $NoneVar$ & 18.44 (83.68) \\
 & $\times$ & $PlainVar$ & 20.26 (81.54) \\ 
 & $\times$ & $PlainExtVar$  & 20.20 (83.38)\\ 
 & $\times$ & $RetroVar$ &  21.25 (81.18) \\
 & $\times$ & $RetroExtVar$  & 21.57 (82.89) \\
  & $\checkmark$ & $NoneVar$ &  22.70 (81.51) \\
 & $\checkmark$ & $PlainVar$ &  19.27 (83.87) \\ 
 & $\checkmark$ & $PlainFixed$ &  21.20 (81.61) \\
 & $\checkmark$ & $PlainExtVar$  & 20.76 (83.17) \\
 & $\checkmark$ & $PlainExtFixed$  & 19.32 (82.38) \\
 & $\checkmark$ & $RetroVar$ &  22.71 (81.12) \\
 & $\checkmark$ & $RetroFixed$ &  \textbf{28.86} (80.53) \\
 & $\checkmark$ & $RetroExtVar$  & 20.95 (81.94) \\
 & $\checkmark$ & $RetroExtFixed$  & \textbf{31.12} (79.63) \\
 \hline
\multirow{6}{*}{\textsc{Copy+SL}} & $\times$ & $NoneVar$ & 17.88 (83.70) \\
& $\times$ & $PlainVar$ & 20.22 (81.52) \\
 & $\times$ & $PlainExtVar$  & 20.14 (83.46) \\   
 & $\times$ & $RetroVar$ &  21.30 (81.22) \\ 
 & $\times$ & $RetroExtVar$  & 21.52 (82.86) \\ 
 & $\checkmark$ & $NoneVar$ &  22.72 (81.41) \\
 & $\checkmark$ & $PlainVar$ &  21.46 (81.39) \\ 
 & $\checkmark$ & $PlainFixed$ &  23.76 (81.68) \\
 & $\checkmark$ & $PlainExtVar$  & 20.68 (83.18) \\
 & $\checkmark$ & $PlainExtFixed$  & 22.23 (81.71) \\
 & $\checkmark$ & $RetroVar$ &  22.62 (81.15) \\ 
 & $\checkmark$ & $RetroFixed$ &  27.66 (81.35) \\
 & $\checkmark$ & $RetroExtVar$  & 24.11 (79.92) \\ 
 & $\checkmark$ & $RetroExtFixed$  & 27.81 (84.67) \\
 \hline 
\end{tabular}
\caption{Test BLEU results. \emph{Sh} denotes encoder-decoder embedding sharing (\textit{No}=$\times$,\textit{Yes}=$\checkmark$) . \emph{Init} denotes the manner of initializing embedding vectors. The \emph{-Fixed} or \emph{-Var} suffix indicates whether embeddings are fixed or trainable. \textsc{COPY} and \textsc{SIMPLES2S} denote presence/absence of \textit{Copy} component. \textsc{+SL} denotes sentinel loss.}
\textbf{\label{tab:knightExp}}
\end{table}


Table \ref{tab:intro} presents model outputs\footnote{All neural outputs are lowercase due to our preprocessing. Although this slightly affects BLEU, it helps prevent token occurrences getting split due to capitalization.} for some test examples. In general, the \textit{Copy} model outputs resemble the ground truth more closely compared to \textit{SimpleS2S} and \textit{Stat} . In some cases, it faces issues with repetition (Examples 4 and 6) and fluency (Example 8).

\section{Related Work} \label{sec:RelatedWord}

There have been some prior work on style adaptation. Xu et al. \shortcite{xu2012paraphrasing} use phrase table based statistical machine translation to transform text to target style.  
On the other hand our method is an end-to-end trainable neural network.
Saha Roy et al \shortcite{saha2015automated} leverage different language models based on geolocation and occupation to align a text to specific style. However, their work is limited to addition of adjectives and adverbs. Our method can handle more generic transformations including addition and deletion of words.

Pointer networks \cite{vinyals2015pointer} allow the use of input-side words directly as output in a neural S2S model, and have been used for tasks like extractive summarization \cite{see2017get} \cite{zeng2016efficient}  and question answering \cite{wang2016machine}. However, pointer networks cannot generate words not present in the input. A mixture model of recurrent neural network and pointer network has been shown to achieve good performance on language modeling task \cite{merity2016pointer}.

S2S neural models, first proposed by \newcite{sutskever2014sequence}, and enhanced with a attention mechanism by \newcite{bahdanau2014neural}, have yielded state-of-the-art results for machine translation (MT),
, summarization \cite{rush2015neural}, etc. In the context of MT, various settings such as multi-source MT \cite{zoph2016multi} and MT with external information \cite{sennrich2016controlling} have been explored. Distinct from all of these, our work attempts to solve a Modern English $\rightarrow$ Shakespearean English style transformation task. Although closely related to both paraphrasing and MT, our task has some differentiating characteristics such as considerable source-target overlap in vocabulary and grammar (unlike MT), and different source and target language (unlike paraphrasing). \newcite{gangal17emnlp} have proposed a neural sequence-to-sequence solution for generating a portmanteau given two English root-words. Though their task also involves large overlap in target and input, they do not employ any special copying mechanism. 
Unlike text simplification and summarization, our task does not involve shortening content length. 

\section{Conclusion} \label{sec:Conclusion}
In this paper we have proposed to use a mixture model of pointer network and LSTM to transform Modern English text to Shakespearean style English.
We demonstrate the effectiveness of our proposed approaches over the baselines. Our experiments reveal the utility of incorporating input-copying mechanism, and using dictionary constraints for problems with shared (but non-identical) source-target sides and sparse parallel data. 

We have demonstrated the transformation to Shakespearean style English only. Methods have to be explored to achieve other stylistic variations corresponding to formality and politeness of text, usage of fancier words and expressions, etc. We release our code publicly to foster further research on stylistic transformations on text. \footnote{https://github.com/harsh19/Shakespearizing-Modern-English}. 




\section*{Acknowledgements}
We thank Taylor Berg-Kirkpatrick and anonymous reviewers for their comments. This research was supported in part by DARPA grant FA8750-12-2-0342 funded under the DEFT program.



\bibliography{emnlp2017}
\bibliographystyle{emnlp_natbib}

\end{document}


\maketitle

\begin{abstract}
    This document contains additional information to complement the descriptions and findings reported in the paper, as well as a guide to the supplementary material.
\end{abstract}

\section{Code}
In the final submission, we will share a link to our github repository for this paper. To not violate double blind submission guidelines, we do not link to it here. 

We do however, provide an (anonymous shareable) Google Drive link\footnote{The file size prevents us from sharing directly as supplmentary material.} of our code, available at \url{http://tinyurl.com/y8z4x25q}. The file is named \textit{TextStyleTransfer.zip}. On unzipping the file, a directory named \textit{TextStyleTransfer/} is created. The root of the directory has suitable running instructions in the form of an included \textit{README.MD}. The directory \textit{code/} contains the code and the directory \textit{data/} contains the data. The code is largely self-contained in terms of data, except that library prerequisites must be met.

\section{Statistical Baselines}
Statistical Machine Translation (SMT) methods uses a noisy-channel formulation to split $P(t|s)$ (distribution over target sequences given source sequence) as
\begin{align*}
    argmax P(t|s) &= argmax P(s|t) P(t)
\end{align*}
$P(s|t)$ is referred to as the alignment model and $P(t)$ as the language model (LM). These models are generally trained independently - the alignment model on the parallel data and the language model on the target side of parallel data (and optionally, additional target-side monolingual data). At test time, each model is represented as a FST and the composition of these FSTs is used for decoding (finding best target sentence). In order to have a finite FST, $P(t)$ has to be a classical N-gram language model. (since neural models such as RNNLMs cannot be represented as FSTs)

In our case, a direct way of incorporating the dictionary $L$ is to add the pairs from $L$ to our parallel data while learning the alignment model $P(s|t)$. A direct way of incorporating the external data (\textit{PTB}), is to use it while training $P(t)$. Another potential enhancement is to use the source side sentences (in addition to target side sentences) while training. We train several models (described below) ablating for the use of each of these enhancements. Models 1-5\footnote{Not to be confused with the canonical IBM SMT models such as IBM Model 1, IBM Model 3 etc. The nomenclature is unrelated.} experiment with the LM while using only the given parallel pairs for training. Models 6-9 use on of the LMs from Model 1-5 while using additional pairs from $L$ while training $P(s|t)$

\begin{enumerate}
    \item \textbf{Model 1}: This model is the simplest. $P(s|t)$ is learnt on parallel sentences and the 3-gram LM $P(t)$ is learnt on target-side training sentences. It does not use $L$ or $PTB$ in any way.
    \item \textbf{Model 2}: This model is same as \textbf{Model 1}, except that it uses a 4-gram LM.
    \item \textbf{Model 3}: Similar to \textbf{Model 2}, except that \textit{PTB} sentences are used while training $P(t)$
    \item \textbf{Model 4}: Similar to \textbf{Model 2}, except that source side sentences are used to train $P(t)$
    \item \textbf{Model 5}: Similar to \textbf{Model 2}, except that both source side sentences and \textit{PTB} sentences are used to train P(t).
    \item \textbf{Model 6}: $P(t)$ is trained similar to \textbf{Model 5}. $L$ is used additionally while training $P(s|t)$.
    \item \textbf{Model 7}: $P(t)$ is trained similar to \textbf{Model 2}, $L$ is used additionally while training. 
    \item \textbf{Model 8}: $P(t)$ is trained similar to \textbf{Model 3}, $L$ is used additionally while training.
    \item \textbf{Model 9}: $P(t)$ is trained similar to \textbf{Model 4}, $L$ is used additionally while training.
\end{enumerate}

\section{Complete Results Table}
\begin{table}
\centering
\scriptsize
\addtolength{\tabcolsep}{-2pt}
\begin{tabular}{|l|l|l|l| }
\hline 
Model & Sh  & Init  & BLEU (PINC) \\ \hline \hline
\textsc{As-it-is}  & {-} & {-}  &  {21.13} (0.0)  \\ \hline
\textsc{Dictionary}  & {-} & {-}  &  {17} (26.64)  \\ \hline
\textsc{Stat} (Model 1)  & {-} & {-}  &  {22.26} (28.34)  \\ \hline
\textsc{Stat} (Model 2)  & {-} & {-}  &  {24.17} (28.14)   \\ \hline
\textsc{Stat} (Model 3)  & {-} & {-}  &  {24.10} (25.86)   \\ \hline
\textsc{Stat} (Model 4)  & {-} & {-}  &  {23.76} (25.86)   \\ \hline
\textsc{Stat} (Model 5)  & {-} & {-}  &  {23.87} (21.85)   \\ \hline
\textsc{Stat} (Model 6)  & {-} & {-}  &  {23.81} (23.60)    \\ \hline
\textsc{Stat} (Model 7)   & {-} & {-}  &  \textbf{24.39} (32.30)    \\ \hline
\textsc{Stat} (Model 8)  & {-} & {-}  &  {24.21} (30.05)    \\ \hline
\textsc{Stat} (Model 9)  & {-} & {-}  &  {23.73} (23.48)   \\ \hline
\multirow{10}{*}{\textsc{SimpleS2S}} &  $\times$ & $NoneVar$ & 11.66 (85.61) \\
&  $\times$ & $PlainVar$ & 9.27 (86.52) \\
 & $\times$ & $PlainExtVar$  & 8.73 (87.17) \\ 
 & $\times$ & $RetroVar$ &  10.57 (85.06) \\ 
& $\times$ & $RetroExtVar$  & 10.26 (83.83) \\ 
& $\checkmark$ & $NoneVar$ &  11.17 (84.91) \\
 & $\checkmark$ & $PlainVar$ &  8.78 (85.57) \\
 & $\checkmark$ & $PlainFixed$ &  8.73 (89.19)\\
 & $\checkmark$ & $PlainExtVar$  & 8.59 (86.04) \\
 & $\checkmark$ & $PlainExtFixed$  & 8.59 (89.16) \\
 & $\checkmark$ & $RetroVar$ &  10.86 (85.58) \\
 & $\checkmark$ & $RetroFixed$ &  11.36 (85.07) \\
 & $\checkmark$ & $RetroExtVar$  & 11.25 (83.56) \\
 & $\checkmark$ & $RetroExtFixed$  & \textbf{10.86} (88.80) \\  \hline
\multirow{6}{*}{\textsc{Copy}} & $\times$ & $NoneVar$ & 18.44 (83.68) \\
 & $\times$ & $PlainVar$ & 20.26 (81.54) \\ 
 & $\times$ & $PlainExtVar$  & 20.20 (83.38)\\ 
 & $\times$ & $RetroVar$ &  21.25 (81.18) \\
 & $\times$ & $RetroExtVar$  & 21.57 (82.89) \\
  & $\checkmark$ & $NoneVar$ &  22.70 (81.51) \\
 & $\checkmark$ & $PlainVar$ &  19.27 (83.87) \\ 
 & $\checkmark$ & $PlainFixed$ &  21.20 (81.61) \\
 & $\checkmark$ & $PlainExtVar$  & 20.76 (83.17) \\
 & $\checkmark$ & $PlainExtFixed$  & 19.32 (82.38) \\
 & $\checkmark$ & $RetroVar$ &  22.71 (81.12) \\
 & $\checkmark$ & $RetroFixed$ &  \textbf{28.86} (80.53) \\
 & $\checkmark$ & $RetroExtVar$  & 20.95 (81.94) \\
 & $\checkmark$ & $RetroExtFixed$  & \textbf{31.12} (79.63) \\
 \hline
\multirow{6}{*}{\textsc{Copy+SL}} & $\times$ & $NoneVar$ & 17.88 (83.70) \\
& $\times$ & $PlainVar$ & 20.22 (81.52) \\
 & $\times$ & $PlainExtVar$  & 20.14 (83.46) \\   
 & $\times$ & $RetroVar$ &  21.30 (81.22) \\ 
 & $\times$ & $RetroExtVar$  & 21.52 (82.86) \\ 
 & $\checkmark$ & $NoneVar$ &  22.72 (81.41) \\
 & $\checkmark$ & $PlainVar$ &  21.46 (81.39) \\ 
 & $\checkmark$ & $PlainFixed$ &  23.76 (81.68) \\
 & $\checkmark$ & $PlainExtVar$  & 20.68 (83.18) \\
 & $\checkmark$ & $PlainExtFixed$  & 22.23 (81.71) \\
 & $\checkmark$ & $RetroVar$ &  22.62 (81.15) \\ 
 & $\checkmark$ & $RetroFixed$ &  27.66 (81.35) \\
 & $\checkmark$ & $RetroExtVar$  & 24.11 (79.92) \\ 
 & $\checkmark$ & $RetroExtFixed$  & 27.81 (84.67) \\
 \hline 
\end{tabular}
\caption{Test BLEU results. \emph{Sh} denotes encoder-decoder embedding sharing. \emph{Init} denotes the manner of initializing embedding vectors. The \emph{-Fixed} or \emph{-Var} suffix indicates whether embeddings are fixed or trainable. \textsc{COPY} and \textsc{SIMPLES2S} denote presence/absence of \textit{Copy} component. \textsc{+SL} denotes sentinel loss.}
\textbf{\label{tab:knightExp}}
\end{table}
